\documentclass[11pt]{article}
\pdfoutput=1

\usepackage{microtype}
\usepackage{graphicx}
\usepackage{subfigure}
\usepackage{booktabs} 
\usepackage{bm}
\usepackage{diagbox}
\usepackage{algorithm,algorithmic}
\usepackage[algo2e,ruled,vlined]{algorithm2e}

\usepackage{geometry}
\usepackage{amssymb}
\usepackage{graphicx}
\usepackage{url}
\usepackage{setspace}
\usepackage[pdftex,bookmarksnumbered,bookmarksopen,
colorlinks,citecolor=blue,linkcolor=blue,urlcolor=blue]{hyperref}
\usepackage{framed}
\usepackage{xcolor}
\usepackage{soul}
\usepackage{longtable}

\usepackage{times}
\usepackage{enumitem}
\usepackage{varwidth}
\usepackage{graphicx}
\usepackage{wrapfig}
\usepackage{enumerate}

\usepackage{amssymb}
\usepackage{graphicx}
\usepackage{url}
\usepackage{setspace}
\usepackage{framed}
\usepackage{xcolor}
\usepackage{soul}
\usepackage{longtable}

\usepackage{times}
\usepackage{enumitem}
\usepackage{varwidth}
\usepackage{graphicx}
\usepackage{wrapfig}
\usepackage{enumerate}

\usepackage[utf8]{inputenc} 
\usepackage[T1]{fontenc}    
\usepackage{url}            
\usepackage{booktabs}       
\usepackage{amsfonts}       
\usepackage{nicefrac}       
\usepackage{microtype}      

\usepackage{mathrsfs}
\usepackage[misc]{ifsym}

\usepackage{graphicx}
\usepackage{appendix}
\usepackage{amsmath,amsfonts,amsthm}
\usepackage{mdwlist}
\usepackage{xspace}
\usepackage{color}
\usepackage{mathrsfs}

\usepackage{booktabs}
\usepackage{comment}

\usepackage{multirow}

\renewcommand{\dim}{\mathrm{dim}}

\newcommand{\sign}{\textup{\textsf{sign}}}

\newcommand{\diag}{\textsf{Diag}}

\newcommand{\adv}{\mathrm{rob}}
\newcommand{\nat}{\mathrm{nat}}

\newcommand{\Appendix}[1]{the full version for}

\newtheorem{theorem}{Theorem}[section]

\newcommand{\x}{\mathbf{x}}
\newcommand{\y}{\bm{y}}
\newcommand{\z}{\mathbf{z}}
\newcommand{\A}{\mathbf{A}}

\newcommand{\D}{\mathbf{D}}

\newcommand{\J}{\mathbf{J}}

\newcommand{\R}{\mathbb{R}}

\newcommand{\W}{\mathbf{W}}

\newcommand{\norm}{\textup{\textsf{norm}}}

\newcommand{\0}{\mathbf{0}}
\newcommand{\1}{\mathbf{1}}
\renewcommand{\comment}[1]{}

\newcommand{\cA}{\mathcal{A}}

\newcommand{\cL}{\mathcal{L}}

\newcommand{\cP}{\mathcal{P}}

\newcommand{\cX}{\mathcal{X}}
\newcommand{\bbB}{\mathbb{B}}

\newcommand{\tabincell}[2]{\begin{tabular}{@{}#1@{}}#2\end{tabular}}

\newcommand{\coau}[1]{{#1$^{\mbox{\tiny{\Letter}}}$}} 	

\title{Adversarial Robustness of Stabilized NeuralODEs Might be from Obfuscated Gradients}
\author{Yifei Huang$^1$, Yaodong Yu$^2$,
\coau{Hongyang Zhang$^{3}$}, Yi Ma$^2$, and \coau{Yuan Yao$^{1}$}\\
~\\
$^1$Department of Mathematics, Hong Kong University of Science and Technology \\
$^2$Department of EECS, University of California at Berkeley \\ 
$^3$University of Waterloo and Toyota Technological Institute at Chicago \\
\coau{} Email: \url{hongyanz@ttic.edu}, \url{yuany@ust.hk}
}
\date{}

\begin{document}

\maketitle

\begin{abstract}%
In this paper we introduce a provably stable architecture for Neural Ordinary Differential Equations (ODEs) which achieves non-trivial adversarial robustness under white-box adversarial attacks even when the network is trained naturally. For most existing defense methods withstanding strong white-box attacks, to improve robustness of neural networks, they need to be trained adversarially, hence have to strike a trade-off between natural accuracy and adversarial robustness. 
Inspired by dynamical system theory, we design a stabilized neural ODE network named SONet whose ODE blocks are skew-symmetric and proved to be input-output stable. With natural training, SONet can achieve comparable robustness with the state-of-the-art adversarial defense methods, without sacrificing natural accuracy. 
Even replacing only the first layer of a ResNet by such a ODE block can exhibit further improvement in robustness, e.g., under PGD-20 ($\ell_\infty=0.031$) attack on CIFAR-10 dataset, it achieves 91.57\% and natural accuracy and 62.35\% robust accuracy, while a counterpart architecture of ResNet trained with TRADES achieves natural and robust accuracy 76.29\% and 45.24\%, respectively. To understand possible reasons behind this surprisingly good result, we further explore the possible mechanism underlying such . We show that the adaptive stepsize numerical ODE solvers, such as adaptive HEUN2, BOSH3, and DOPRI5, have a gradient masking effect that fails the PGD attacks which are sensitive to gradient information of training loss; on the other hand, they cannot fool the CW attack of robust gradients and the SPSA attack that is gradient-free. This provides a new explanation that the adversarial robustness of ODE-based networks mainly comes from the obfuscated gradients in numerical ODE solvers with adaptive step sizes. (Source codes: \url{https://github.com/silkylove/SONet}; \url{https://github.com/yao-lab/SONet})
\end{abstract}

\section{Introduction}
Adversarial robustness is a central object of study in machine learning~\cite{carlini2019evaluating,zhang2019deep,madry2018towards,kolter2017provable}, computer security~\cite{sharif2016accessorize,meng2017magnet}, and many other domains~\cite{song2018pixeldefend,xie2017adversarial,jia2017adversarial}. In machine learning, study of adversarial robustness has led to significant advance in understanding the generalization~\cite{schmidt2018adversarially,carmon2019unlabeled,alayrac2019labels,zhai2019adversarially}, interpretability of learning models~\cite{tsipras2018robustness}, and connecting robust statistics~\cite{gao2018robust,gao2019generative}. In computer security, adversarial robustness serves as an indispensable component towards AI safety against adversarial threat, in a range of security-critical systems and applications such as autonomous vehicles~\cite{eykholt2018robust} and biometric authorization~\cite{thys2019fooling}. The problem of achieving adversarial robustness can be stated as learning a classifier with high test accuracy on both natural and \emph{adversarial examples}. The adversarial example is either in the form of unrestricted transformations, such as rotation and translation of natural examples, or in the form of perturbations with bounded norms. The focus of this work is the latter setting.
 
Probably one of the most successful techniques to enhance model robustness is by \emph{adversarial training}~\cite{madry2018towards,zhang2019theoretically}. In the adversarial training, the defenders simulate adversarial examples against current iteration of model and then feed them into the training procedure in the next round. Despite a large literature devoted to the study of adversarial training, many fundamental questions remain unresolved. One of the long-standing questions is the interpretability: although adversarial training is an effective way to defend against certain adversarial examples, it remains unclear why current designs of network architecture are vulnerable to adversarial attacks without adversarial training. This question becomes more challenging when we consider the computational issues. Taking the perspective of Pontryagin Maximum Principle (or Bellman Equation) for differential games induced by adversarial training, \cite{yiping2019_yopo} reduces adversarial training to merely updating the weights of the first layer that significantly reduces the computational cost. Yet in optimization, adversarial training is notorious for its instability due to the non-convex non-concave minimax nature of its loss function.

When the ``simulated'' adversarial examples in the training procedure do not conceptually match those of attackers, adversarial training can be vulnerable to adversarial threat as well. This is known as the norm-agnostic setting, and there is significant evidence to indicate that adversarial training suffers from brittleness against attacks in $\ell_2$ and $\ell_\infty$ norms simultaneously~\cite{li2019norm}. Furthermore, due to an intrinsic trade-off between natural accuracy and adversarial robustness~\cite{tsipras2018robustness,zhang2019theoretically}, adversarial training typically leads to more than $10\%$ reduction of accuracy compared with natural training.

Stability principle of dynamical systems has been applied to adversarial training to enhance the robustness. Inspired by the initial value stability of convection-diffusion partial differential equation and the Feynman-Kac formula of solutions, \cite{osher2019feynmankac} designs ResNet ensembles with activation noise that exhibits improvements in both natural and robust accuracies for adversarial training. Moreover, motivated by the fact in numerical ODEs that implicit (backward) Euler discretization has better stability than explicit (forward) Euler discretization that current ResNets exploit, \cite{zhouchen2020ieuler} designs implicit Euler based skip-connections to enhance ResNets with better stability and adversarial robustness.  However, all these studies are limited to adversarial training rather than natural training. 

In response to the limitations of adversarial training, designing network architecture towards natural training as robust as adversarial training has received significant attention in recent years. On one hand, most positive results for obtaining adversarial robustness have focused on controlling Lipschitz constants explicitly in the training procedure, such as requiring each convolutional layer be composed of orthonormal filters~\cite{cisse2017parseval}, or restricting the spectral radius of the matrix in each layer to be small~\cite{qian2018lnonexpansive}. These approaches, however, do not achieve comparable robustness as adversarial training against $\ell_\infty$-norm attacks. 
On the other hand, with the introduction of ordinary different equations into neural networks~\cite{chen2018neural}, the adversarial robustness for neural ordinary differential equations (ODEs) network architecture have been attracting rising attention. \cite{yan2019robustness} found that ODE networks with natural training is more robust against adversarial examples compared with traditional conventional neural networks, but the robustness of ODE networks is much weaker than the state-of-the-art result by adversarial training.

\subsection{Our methodology and results}

We begin with designing ODE networks analogous to the residual networks. Our ODE network is a natural extension of the Residual Network: when we solve the ODE system by the explicit (forward) Euler method, the two types of networks can be made equivalent. Nevertheless, to ensure the output of our networks to be less sensitive to perturbations in input, we further require our ODE networks to be stable by design. It has been well known in the dynamical system theory \cite{Cal:Des} that input-output stability is an important property for a system to be insensitive (and even robust) to input noise and perturbations. We rigorously show that the resulting networks are stable in the Lyapunov sense, provided that the two weight matrices in each ODE block are skew-symmetric to each other up to an arbitrarily small damping and the activation function is strictly monotonically increasing. The design works for both convolutional and fully-connected neural networks.

Our stability analysis naturally leads to a new formulation of network architecture
which has several appealing properties; in particular, it inherits all the benefits of Neural ODE such as parameter- and memory-efficiency, adaptive computation, etc., and the algorithm achieves comparable robustness on a range of benchmarks as the state-of-the-art adversarial training methods. To understand possible reasons behind this surprisingly good result, we further explore possible mechanisms and disclose the obfuscated gradients caused by adaptive step sizes of numerical ODE solvers.

The main contribution and discovery in this report can be summarized as follows.


\begin{itemize}
    \item
    Theoretically, we parametrize ODE networks analogous to the residual networks. Our stability analysis shows that the ODE system is Lyapunov stable, provided that the activation function is strictly monotonically increasing and the two weight matrices in the ODE block are skew-symmetric with each other, up to an arbitrarily small damping factor.
    \item
    Algorithmically, inspired by our stability analysis, we propose a new formulation of neural ODE network architecture, named \textbf{S}tabilized neural \textbf{O}DE \textbf{Net}work (SONet). The architecture is robust to small perturbations as each ODE block is provably stable in the sense of Lyapunov.
    \item
    Experimentally, we show that natural training of the proposed architecture achieves non-trivial adversarial robustness in white-box PGD attacks, and even better than the state-of-the-art ResNet10 adversarially trained by TRADES under white-box $\ell_\infty$ and $\ell_2$ PGD$^{20}$ attacks. 
    \item Furthermore, a possible interpretation for the adversarial robustness of ODE-based networks is provided, suggesting that numerical ODE solvers with adaptive step sizes (e.g. adaptive HEUN2, BOSH3, and DOPRI5) may lead to obfuscated gradients via large error tolerance in adaptive step size choice, which fails the gradient based attacks like PGD but may not fool robust gradient attacks like CW and gradient-free attacks like SPSA. 
\end{itemize}

\section{Introduction}

Before proceeding, we define some notations and formalize our model setup in this section.

\subsection{Notations}

\begin{wrapfigure}{R}{6.5cm}
\subfigure[ResNet block]{
\includegraphics[width=3cm]{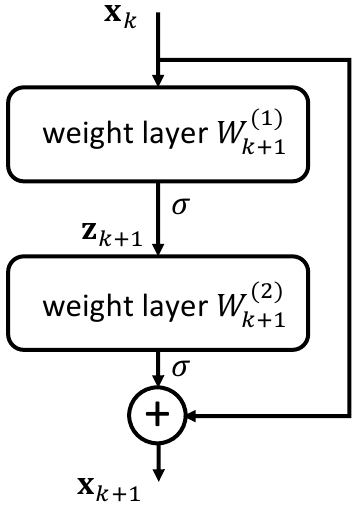}\label{figure: network architecture a}
}
\subfigure[ODENet block]{
\includegraphics[width=3cm]{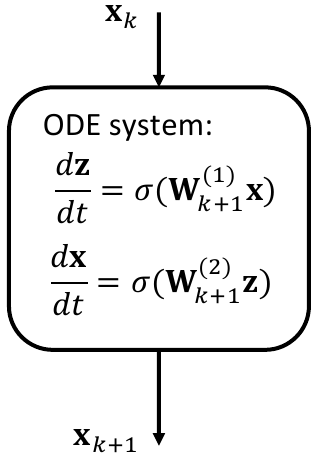}\label{figure: network architecture b}
}
\vspace{-0.5cm}
\caption{Network architecture.}
\vspace{-0.2cm}
\end{wrapfigure}

We will use bold capital letters such as $\W$ to represent matrices, bold lower-case letters such as $\x$ to represent vectors,
and lower-case letters such as $t$ to represent scalars. Specifically, we denote by $\0$ the all-zero vector, by $\1$ the all-one vector, by $\x\in\R^{d_\text{in}}$ the input vector to each architecture block, and by $\z\in\R^{d_\text{out}}$ the output vector, where $d_\text{in}$ does not necessarily equal to $d_\text{out}$. Denote by $\sigma(\cdot)$ the element-wise activation function, and $\sigma'(\cdot)$ is its (sub-)gradient. We will frequently use $d\x(t)/dt$ to represent the differential of $\x(t)$ w.r.t. the time variable $t$. For norms, we denote by $\|\cdot\|$ a generic norm. Examples of norms include $\|\x\|_p$, the $\ell_p$ norm of vector $\x$ for $p\ge 1$. We will use $f_1\circ f_2(\cdot)$ to represent the composition of two functions $f_1(\cdot)$ and $f_2(\cdot)$. Denote by $\mathbb{B}(\x,\epsilon)$ a neighborhood of $\x$: $\{\x': \|\x-\x'\|\le \epsilon\}$. Throughout the paper, for any given loss function $\cL(f,\x)$ and data (set) $\x$, we will term the optimization procedure $\min_f \max_{\x'\in \mathbb{B}(\x,\epsilon)} \cL(f,\x')$ as \emph{adversarial training} and term the optimization procedure $\min_f \cL(f,\x)$ as \emph{natural training}.

\subsection{ODE Blocks}
In the Residual Networks (ResNets, a.k.a. Euler networks)~\cite{he2016deep}, the basic blocks follow the architecture\footnote{Without loss of generality, we assume the bias term is a zero vector for simplicity, although our analysis works for the general case as well.} (see Figure \ref{figure: network architecture a}):
\begin{align}\label{equ: ResNet block}
\frac{\z_{k+1}-\z_k}{ \Delta t}&=\sigma(\W_{k+1}^{(1)}\x_k),\nonumber\\
\frac{\x_{k+1}-\x_k}{ \Delta t}&=\sigma(\W_{k+1}^{(2)}\z_{k+1}),\\
\z_k&=0,\ \Delta t=1,\nonumber
\end{align}
where $\x_k$ and $\x_{k+1}$ are the input and ouput of the $k$-th ResNet block, $\z_{k+1}$ is the intermediate layer, and $\W_{k+1}^{(1)}$ and $\W_{k+1}^{(2)}$ are the weight matrices which represent either the fully-connected or the convolutional operators.
In the Neural ODE, in constrast, \cite{chen2018neural} took the limit of the finite differences over the infinitesimal $\Delta t$ and parameterized the continuous dynamics of hidden units using an ODE specified by a neural network:
\begin{align}\label{equ:neural-ODE}
\x_{k+1}=f_{\text{NeuralODE-k}}(\x_k; t_0):\quad \frac{d\x(t)}{dt}=\sigma(\W_{k+1}^{(2)}\sigma(\W_{k+1}^{(1)}\x(t))),\ \x(0)=\x_k,\ \x_{k+1}=\x(t_0),
\end{align}
where $\x_k$ is the initial condition of $\x(t)$, i.e., the input, and the output $\x_{k+1}$ is the evolution of $\x(t)$ at time $t_0$.

Our study is motivated by the Neural ODE. We focus on a parametric model similar to ResNet block \eqref{equ: ResNet block} (see Figure \ref{figure: network architecture b}):
\begin{align}
\label{equ: stable ODE block}
\frac{d\z(t)}{dt}&=\sigma(\W_{k+1}^{(1)}\x(t) - \gamma \z(t)),\nonumber \\
\frac{d\x(t)}{dt}&=\sigma(\W_{k+1}^{(2)}\z(t) - \gamma \x(t)),\\ 
\x_{k+1}&=\z(t_0), \quad \x(0)=\x_k,\quad  \z(0)=\z_k,\nonumber
\end{align}
where $\x_k$ and $\z_k$ are the initial conditions of $\x(t)$ and $\z(t)$, respectively, $\gamma>0$ is a small positive constant as the damping factor and the output $\x_{k+1}$ is the evolution of $\z(t)$ at time $t_0$. When we solve ODE system \eqref{equ: stable ODE block} by the Euler solver with time step $1$ and set $\gamma$ to be 0, ODE block \eqref{equ: stable ODE block} is equivalent to ResNet block \eqref{equ: ResNet block}.

\medskip
\noindent{\textbf{Flexibility of parametric model in \eqref{equ: stable ODE block system}.}} Compared with the previous Neural ODE~\cite{chen2018neural} defined in \eqref{equ:neural-ODE}, which can only deal with the case when the size of input $\x_{k}$ is equal to the size of  output $\x_{k+1}$, the parametric model in \eqref{equ: stable ODE block} is able to handle the case when $\dim(\x_{k+1}) \neq \dim(\x_{k})$. The intermediate layer $\z(t)$ can be viewed as an auxiliary layer, which makes our model more flexible.

\section{Stability of ODE Blocks}

In this section, we present our stability analysis for ODE system \eqref{equ: stable ODE block} that serves as a guiding principle in the design of network architecture against adversarial examples.
Our analysis leads to the following guarantee on the stability of the ODE system.
\begin{theorem}[Stability of ODE Blocks]\label{theorem: stability of ODE block}
Suppose that the activation function 
$\sigma$ is strictly monotonically increasing, i.e.,
$\sigma^{\prime}(\cdot)>0$ 
and positive damping factor $\gamma$ is small. 
Let $\W_{k+1}^{(2)}=-\W_{k+1}^{(1)\top}$. 
Then for any implementation of network parameters, the forward propagation \eqref{equ: stable ODE block} is stable in the sense of Lyapunov; 
that is, for all $\delta>0$, there exists a stable radius $\epsilon(\delta)>0$ such that if $\|\x_0-\x_0'\|\le \epsilon(\delta)$, we have $\| f_\textup{ODENet-k}(\x_0; t_0)-f_\textup{ODENet-k}(\x_0'; t_0)\|\le \delta$ for all $t_0>0$. 
\end{theorem}

Theorem \ref{theorem: stability of ODE block} demonstrates that there exists a \emph{universal} stability radius $\epsilon>0$ (independent of integration time $t_0$) such that small change of $\x_0$ within the $\epsilon$-ball causes small change of $f_\textup{ODENet-k}(\x_0;t_0)$ for \emph{all $t_0>0$}. In contrast, the continuity in the original design of Neural ODE~\cite{chen2018neural} does not justify the existence of such universal stability radius for all $t_0>0$. Our theory shows that the quantity $\| f_\textup{ODENet}(\x_0; t)-f_\textup{ODENet}(\x_0'; t_0)\|$ does not diverge as $t_0$ grows.
So the ODE is robust w.r.t. its initial condition, the input of the network.



\medskip
\noindent{\textbf{Intuition behind the stability.}}
Our ODE block \eqref{equ: stable ODE block} has guaranteed stability without any explicit regularization on the smoothness of its input and output. To see how the skew-symmetric architecture encourages stability, let us ignore for now the nonlinear activation $\sigma(\cdot)$ and $\gamma$ in the ODE block \eqref{equ: stable ODE block} and consider its linearized version:
\begin{align}
\label{equ: stable ODE block system}
\frac{d}{dt}
\begin{bmatrix}
\x\\
\z
\end{bmatrix}
&=
\begin{bmatrix}
\0 & \W_{k+1}^{(2)}\\
-\W_{k+1}^{(2)\top} & \0
\end{bmatrix}
\begin{bmatrix}
\x\\
\z
\end{bmatrix} \doteq \mathbf A_{k+1} \begin{bmatrix}
\x\\
\z
\end{bmatrix},\\
\ \x_{k+1}&=\z(t), \quad \x(0)=\x_k,\quad \z(0)=\z_k.\nonumber
\end{align}
As the linear system matrix $\mathbf A_{k+1}$ is skew symmetric, one can show that the solution to the above system is given by \cite{Cal:Des}: 
\begin{equation*}
    \begin{bmatrix}
\x(t)\\
\z(t)
\end{bmatrix} = \mathbf \Phi \begin{bmatrix}
\x(0)\\
\z(0)
\end{bmatrix},
\end{equation*}
where the state-transition matrix $\mathbf{\Phi}$ is an orthogonal matrix $\mathbf{\Phi} \mathbf{\Phi}^{\top} = \mathbf{I}$. Hence the input-output of the linearized system is always norm-preserving. 

Below we give a formal proof of stability of the ODE block with the nonlinear activation, i.e. Theorem \ref{theorem: stability of ODE block}, based on results from system theory \cite{astrom2010feedback}.

\medskip

\begin{proof}
We observe that Eqn. \eqref{equ: stable ODE block} has an equivalent expression, $\x_{k+1}=f_{\text{ODENet}}(\x_k;t_0)$:
\begin{align*}
    \frac{d}{dt}
\begin{bmatrix}
\x\\
\z
\end{bmatrix}
=\sigma \left(
\begin{bmatrix}
\0 & -\W_{k+1}^\top\\
\W_{k+1} & \0
\end{bmatrix}
\begin{bmatrix}
\x\\
\z
\end{bmatrix}-\gamma \mathbf{I}
\begin{bmatrix}
\x\\
\z
\end{bmatrix}
\right),\\
\ \x(0)=\x_k,\ \z(0)=\z_k, \ \x_{k+1}:=\z(t_0).
\end{align*}
Denote by $$\A_{k+1}:=
\begin{bmatrix}
\0 & -\W_{k+1}^\top\\
\W_{k+1} & \0
\end{bmatrix}.$$
Note that $\A_{k+1}$ is a skew-symmetric matrix such that $\A_{k+1}=-\A_{k+1}^{\top}$. So $Re[\lambda_i(\A_{k+1})]\le 0$ for all $i$, where $Re[\cdot]$ represents the real part of a complex variable and $\lambda_i(\A_{k+1})$ is the $i$-th eigenvalue of matrix $\A_{k+1}$.

We note that an ODE system is stable if $Re[\lambda_i(\J_{k+1})]< 0$~\cite{astrom2010feedback}, where $\J_{k+1}$ is the Jacobian of the ODE:
\begin{equation*}
\begin{split}
\J_{k+1}&:=\nabla_{[\x;\z]}\left(\sigma \left(
\begin{bmatrix}
\0 & -\W_{k+1}^\top\\
\W_{k+1} & \0
\end{bmatrix}
\begin{bmatrix}
\x\\
\z
\end{bmatrix}-
\gamma \mathbf{I}
\begin{bmatrix}
\x\\
\z
\end{bmatrix}
\right)\right)\\
&=:\D_{k+1}(\A_{k+1} - \gamma \mathbf{I}),
\end{split}
\end{equation*}
where we have defined
\begin{equation*}
\D_{k+1}:=
\diag\left(\sigma' \left(
\begin{bmatrix}
-\gamma & -\W_{k+1}^\top\\
\W_{k+1} & -\gamma
\end{bmatrix}
\begin{bmatrix}
\x\\
\z
\end{bmatrix}
\right)\right).
\end{equation*}
Because $\sigma'(\cdot)>0$, the matrix $\D_{k+1}^{-1/2}$ exists. We observe that
\begin{equation*}
\J_{k+1}\sim \D_{k+1}^{1/2}(\A_{k+1}-\gamma\mathbf{I})\D_{k+1}^{1/2},
\end{equation*}
where the notation $\sim$ means the two matrices are similar. Since similar matrices have the same eigenvalues, for all $i$, we have
\begin{equation}
\label{equ: same eigenvalue}
\lambda_i(\J_{k+1})=\lambda_i(\D_{k+1}^{1/2}(\A_{k+1}-\gamma\mathbf{I})\D_{k+1}^{1/2}).
\end{equation}
For the right hand side in Eqn. \eqref{equ: same eigenvalue},
$Re[\lambda_i(\A_{k+1})]\le 0$
So $Re[\lambda_i(\A_{k+1}-\gamma\mathbf{I})]< 0$, and matrix $\D_{k+1}$ is positive diagonal. Combining with Eqn. \eqref{equ: same eigenvalue}, we have $Re[\lambda_i(\J_{k+1})]< 0$. Thus, we have $\|(\x(t),\z(t))\|\leq\|(\x(0),\z(0))\|$ and when we set the initial condition $\z(0)=\z(k)=\x(k)$, there holds $\|\x(t)\|\leq\|(\x(t),\z(t))\|\leq\|(\x(0),\z(0))\|\leq \sqrt{2} \| \x(0) \|$ for any $t>t_0$. Alternatively, one can also achieve $\|\x(t)\|\leq\|(\x(t),\z(t))\|\leq\|(\x(0),\z(0))\|= \| \x(0) \|$ if we choose initialization $\z(0)=0$. Finally, the Lyapunov stability is valid with respect to Euclidean $\ell_2$-norm. For other equivalent $\ell_p$-norms ($1\leq p \leq \infty$), the result holds up to a constant that depends on the input dimension. The proof is completed.
\end{proof}

Another quantity governing the robustness of a network is its depth. Empirically, deeper networks enjoy better robustness against adversarial perturbations~\cite{madry2018towards}.
This is probably because the score function of a ReLU-activated neural network is characterized by a piecewise affine function~\cite{croce2018provable}; deeper neural network implies smoother approximation of the ground-truth score function.
Since ODE networks are provable deep limit of ResNets~\cite{avelin2019neural,thorpe2018deep}, the proposed networks implicitly enjoy the benefits of depth.

\section{Architecture Design of ODE Networks}

\medskip
\noindent{\textbf{Architecture design of ODE blocks.}}
Theorem \ref{theorem: stability of ODE block} sheds light on architecture designs of ODE blocks. In order for the ODE to be stable w.r.t. its input at the inference time, the theorem suggests parametrizing ODE network \eqref{equ: stable ODE block} with $\W_{k+1}^{(2)}=-\W_{k+1}^{(1)T}$ and a strictly increasing activation function. We name our network SONet, standing for {\bf S}tabilized {\bf O}DE {\bf Net}work.

Probably the most relevant work to our design is that of \cite{haber2017stable}, where \cite{haber2017stable} proposed similar skew-symmetric architecture, but for the Euler networks. In addition, \cite{haber2017stable} discussed the proposed architecture in the context of exploding and vanishing gradient phenomenon. In contrast, our work sheds light on algorithmic designs for adversarial defenses which is different to \cite{haber2017stable}. We show that a good ODE solver for problem \eqref{equ: stable ODE block} suffices to imply a robust network to adversarial attacks.

\begin{figure*}[ht]
  \begin{center}
    \includegraphics[width=0.99\textwidth]{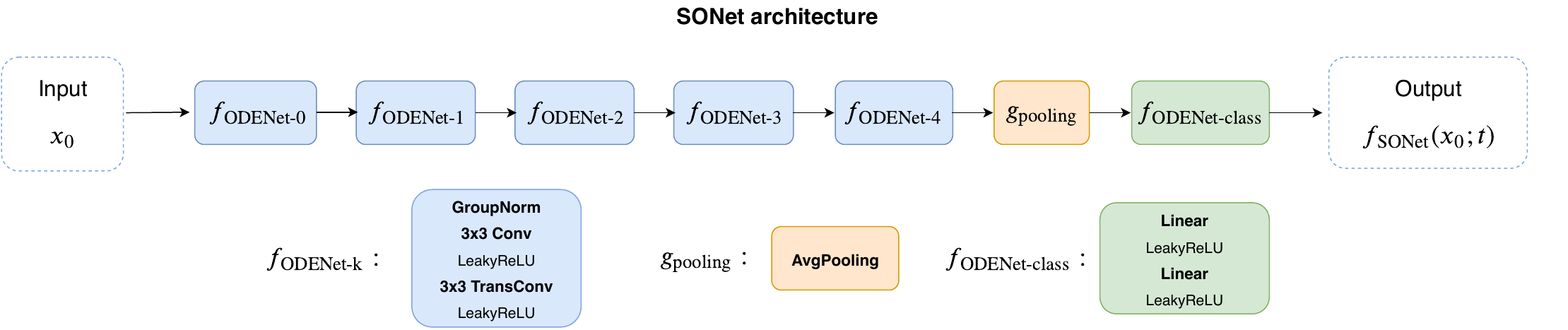}
    \caption{Stabilized neural ODE Network (SONet) architecture example. Both $f_{\text{ODENet-k}}$ and $f_{\text{ODENet-class}}$ are built on our stable ODE block defined in \eqref{equ: stable ODE block}.}
    \label{fig:arch}
  \end{center}
  \vskip -0.1in
\end{figure*}

\medskip
\noindent{\textbf{Benefits of skew-symmetric architecture.}}
The skew-symmetric architecture of ODE blocks has many structural benefits that one can exploit. \emph{Change of dimensionality}: the introduction of the auxiliary variable $\z\in\R^{d_\text{out}}$ enables us to change the dimension of the input and output vectors; that is, the input variable $\x\in\R^{d_\text{in}}$ may have different dimensions as the output variable $\z\in\R^{d_\text{out}}$. This is in sharp contrast to the original design of Neural ODE~\cite{chen2018neural}, where the input and output vectors of each ODE block must have the same dimension. \emph{Parameter efficiency}: the skew-symmetric ODE block has only half number of parameters compared to the ResNet blocks and the original design of Neural ODE blocks due to parameter sharing. \emph{Inference-time robustness}: the established architecture enjoys stability (see Theorem \ref{theorem: stability of ODE block}). Furthermore, an expected side-benefit of our design is that it automatically inherits all the benefits of Neural ODE~\cite{chen2018neural}: memory efficiency, adaptive computation, invertible normalizing flows, and many others.

\medskip
\noindent{\textbf{Construction of robust ODE networks.}} Our construction of ODE networks builds upon the architecture design of ODE block \eqref{equ: stable ODE block}.

\begin{itemize}
\item \emph{Feature-extraction block} $f_\text{ODENet-k}$: The feature blocks aim at extracting the feature of each instance. For the image classification tasks, the operation of multiplying $\z$ with $\W_{k+1}$ in Eqn. \eqref{equ: stable ODE block} serves as a convolution operator, and correspondingly, $\W_{k+1}^T$ serves as a \emph{transposed convolution} (a.k.a. de-convolution) operator which shares a common kernel with $\W_{k+1}$. The input and output dimensions $d_\text{in}$ and $d_\text{out}$ are equal. We set the initial condition $\z_k$ as $\x_k$, the input of the ODE block.
\item \emph{Classification block} $f_\text{class}$: At the top layer of the network is the classification block which is characterized by a fully-connected operator, mapping the extracted feature to the confidence value associated with each class. The matrix $\W_{k+1}$ parametrizes the weight matrix of the fully-connected layer and $\W_{k+1}^T$ is its transpose. The input dimension $d_\text{in}$ and the output dimension $d_\text{out}$ are equal to the feature size and the number of classes, respectively, so $d_\text{in}$ might not equal to $d_\text{out}$. We set the initial condition $\z_k=\z(0)$ as $\1$. This is conceptually consistent with the argument that we have no prior knowledge on the true label of any given instance.
\end{itemize}
Our ODE network is therefore a stack of various building blocks:
\begin{equation}
\label{equ: ODE network}
    f_\text{ODENet}(\x_0; t):=f_\text{class}\circ g_\text{pooling}\circ f_\text{ODENet-L}\circ \cdots \circ f_\text{ODENet-0}\circ g_\text{channel-copy}(\x_0),
\end{equation}
where $\x_0$ is the input instance, $L$ is the number of layers, $g_\text{pooling}$ represents the average pooling operator, and $g_\text{channel-copy}$ is the ``channel-copy'' layer which copies $\x_0$ along the channel direction in order to increase the width of the network. The function $f_\text{ODENet}:\cX\rightarrow \R^C$ is the \emph{score function} which
maps an instance to logits over classes. An example network is shown in Figure~\ref{fig:arch}, this example consists 5 feature-extraction blocks, a pooling layer and a classification block.

\medskip
\noindent{\textbf{Comparisons with prior work.}} We compare our approach with several related lines of research in the prior
work. One of the best known algorithms for adversarial robustness is based on adversarial training. The algorithms approximately solve a minimax problem
\begin{equation*}
    \min_{f} \sum_{i=1}^n \left\{\max_{\x_{(i)}'\in \mathbb{B}(\x_{(i)},\epsilon)} \cP(f,\x_{(i)}')\right\},
\end{equation*}
where $\x_{(i)}$ represents the $i$-th instance, and $\cP(\cdot,\cdot)$ is the payoff function between defender and attacker, which captures the smoothness of model $f$ in an explicit manner; examples of $\cP(f,\x_{(i)}')$ include robust optimization $\cL(f(\x_{(i)}'),\y_{(i)})$~\cite{madry2018towards} and TRADES $\cL(f(\x_{(i)}'),\y_{(i)})+\beta\cL(f(\x_{(i)}'),f(\x_{(i)}))$~\cite{zhang2019theoretically}, where $\cL$ is the cross-entropy loss and $\y_{(i)}$ is the one-hot label. Prior to ours, random smoothing~\cite{cohen2019certified} and stability training~\cite{zheng2016improving} are other techniques towards natural training as robust as adversarial training by adding small Gaussian noises to the input images. Our work, on the other hand, is paralleled to these two lines of research as we focus on the network architecture design. The combination of these methods may further enhance the robustness of learning systems.

Another more related line of research is by network architecture design. Parseval networks~\cite{cisse2017parseval} and $L_2$-nonexpansive neural networks~\cite{qian2018lnonexpansive} explicitly bounded the Lipschitz constant by either requiring each fully-connected or convolutional layer be composed of orthonormal filters~\cite{cisse2017parseval}, or restricting the spectral radius of the matrix in each layer to be small~\cite{qian2018lnonexpansive}. In complex problem domains, however, the explicit Lipschitz constraints may negatively affect the expressive power of the networks and overly trade off natural accuracy against adversarial robustness.~\cite{xie2019feature} involved non-local mean denoiser in the architecture design of ResNet. But the model is vulnerable to attcks without adversarial training. \cite{svoboda2018peernets} proposed PeerNets, a
family of convolutional networks alternating classical Euclidean convolutions with graph convolutions. \cite{yan2019robustness} explored
robustness properties of neural ODEs and proposed the time-invariant steady neural ODE (TisODE), which regularizes
the flow on perturbed data. But the model is weak under PGD attacks. In contrast, in this work, we explore the possibility of enhancing robustness for classic (non-graph) networks with natural training.

\section{Adversarial Robustness under PGD Attacks}
In this section, we evaluate the adversarial robustness of our proposed architecture against projected gradient descent attacks and show the Stablized ODE Block/Net can achieve the state-of-the-art performance even competitive to adversarial training by TRADES, without losing natural accuracy. We use $\cA_\nat(f)$ to denote the natural accuracy of the model, and  $\cA_\adv(f) $ to denote the robust accuracy against adversarial attacks. Additional experiments are provided in Appendix.

\subsection{Experimental Setup}
We first introduce the experimental setup for datasets, deep neural network architectures, adversarial attacks and adversarial defense methods.

\textbf{ResNet}: We apply the ResNet with 10 layers as the baseline model, denoted by ResNet10. Compared with ResNet18 with 2-layer basic block, we use 1-layer basic block for ResNet10. The first layer of ResNet10 is the convolution layer with Batchnorm~\cite{ioffe2015batch} and ReLU activation, then followed by four 1-layer residual basic blocks. Within each basic block there are two convolution layers. Average pooling is applied after the residual blocks and the last layer is a fully connected layer with softmax.

\textbf{SONet}: We apply our stable skew-symmetric ODE block (defined in Eqn. \eqref{equ: stable ODE block}) as the building block in the SONet. More specifically, we replace each residual basic block in the ResNet10 architecture with the proposed stable skew-symmetric ODE block. Besides the replaced residual blocks, as shown in Figure~\ref{fig:arch}, we replace the same first convolution layer with the stable ODE block $f_{\text{ODENet-0}}$, and replace the last fully connected linear layer with the stable ODE block $f_{\text{ODENet-class}}$. 

\textbf{SOBlock}: We replace the first convolution layer in \textbf{ResNet10} above by our proposed skew-symmetric ODE block (defined in Eqn. \eqref{equ: stable ODE block}) and leave the other parts unchanged.

Additionally, in order to compare the performance of SONet, SOBlock and ResNet with different number of parameters, we scale the model capacity by changing the input channel from 32 to 64.

\textbf{Adversarial attacks}: For \textbf{White-Box} attacks, we focus on $\ell_{\infty}$-norm, $\ell_{2}$-norm projected gradient descent (PGD) and CW$_\infty$~\cite{carlini2017towards} adversairal attacks to evaluate the adversarial robustness of different models. For $\ell_{\infty}$ PGD attack, the update rule is defined as
$\x_i'\leftarrow \Pi_{\bbB_{\infty}(\x_i,\epsilon)}(\x_i'+\eta_1\,\sign(\nabla_{\x_i'} \cL(f(\x_{(i)}'),\y_{(i)}))),$
where $\Pi_{\bbB_{\infty}(\cdot,\cdot)}$ is the projection operator with respect to $\ell_{\infty}$-norm, $\cL$ is the cross-entropy loss, $\x_i'$ is initialized as the original input $\x_i$, $\epsilon$ is the adversarial perturbation distance, and $\eta_1$ is the attack step size. For $\ell_{2}$ PGD attack, the update rule is defined as
$\x_i'\leftarrow \Pi_{\bbB_{2}(\x_i,\epsilon)}(\x_i'+\eta_1\,\norm_{2}(\nabla_{\x_i'} \cL(f(\x_{(i)}'),\y_{(i)}))),$
where $\Pi_{\bbB_{2}(\cdot,\cdot)}$ is the projection operator with respect to $\ell_{2}$-norm, and the $\norm_{2}$ is the normalization operator, i.e., $\norm_{2}(\x) = \x/\|\x\|_{2}$. For CW$_\infty$ attack, it minimizes $c \cdot f(x+\delta)+\|\delta\|_{\infty}$ with respect to $\delta$ such that $x+\delta \in[0,1]^{n}$ where $c > 0$ is a suitably chosen constant. For \textbf{Black-Box} attack, we use simultaneous perturbation stochastic approximation (SPSA)~\cite{uesato2018adversarial} adversarial attack which is one of the most powerful gradient free attacks and it minimizes $m_{\theta}(x)_{y_{0}}-\max _{j \neq y_{0}} m_{\theta}(x)_{j}$ with respect to $x$ such that $\left\|x-x_{0}\right\|_{\infty}<\epsilon$ where $m_{\theta}(x)_{j}$ denotes the output logit for the class $j$ and $y_0$ is the true label.
Unless explicitly stated, on CIFAR10 dataset, we set the  perturbation distance $\epsilon=0.031$, the attack step size $\eta_1=0.003$ for $\ell_{\infty}$ PGD attack, and we set the perturbation distance $\epsilon=0.5$, the attack step size $\eta_1=0.1$ for $\ell_{2}$ PGD attack. For CW$_\infty$ attack, we set the perturbation distance $\epsilon=0.031$, the max-iterations $K=100$. And we apply the $\epsilon=0.031$, the number of iterations $K=20$ and the number of samples to be 32 for SPSA attack.

\textbf{Adversarial training}: We use  TRADES~\cite{zhang2019theoretically} as our adversarial training baselines for comparison. We do not compare with other adversarial training approaches because TRADES is known as the state-of-the-art defense method which won the NeurIPS 2018 Adversarial Vision Challange~\cite{brendel2020adversarial}. On CIFAR10 dataset, we set the $\ell_\infty$ perturbation distance $\epsilon=0.031$, perturbation step size $\eta_{1}=0.007$, number of iterations $K=10$ for TRADES. For TRADES, we train two models and set the  regularization parameter as $1/\lambda = 1.0$ and $1/\lambda = 6.0$. 

\textbf{Training settings}: 
On CIFAR10 dataset, for all mentioned models, we set the total epoch $T =350$, batch size $B= 100$, the initial learning rate $\eta= 0.01$ (decay 0.1 at 150 and 300 epochs respectively), and apply stochastic gradient descent (SGD) with momentum 0.9 as the optimizer. No weight decay is used during training. Unless explicitly stated, for all skew-symmetric ODE blocks, we set the constant $\gamma$ in \eqref{equ: stable ODE block} to be 0 and use DOPRI5 solver which is an adaptive solver with 0.1 error tolerance.


\subsection{Projected Gradient Descent (PGD) Attacks}

We shall see that our proposed network, SONet, is able to achieve nontrivial $\ell_{2}$ and $\ell_{\infty}$ white-box PGD robust accuracy on CIFAR10 dataset, only with natrual training. Moreover, SOBlock even outperforms ResNet10 with TRADES training with regard to both natural accuracy and robust accuracy on both PGD$^{20}$ and PGD$^{1000}$ attacks, although TRADES is able to achieve a better tradeoff compared with standard adversarial training~\cite{zhang2019theoretically}.

\subsubsection{Better robustness of SONet in natural training than TRADES adv-training in PGD$^{20}$ attacks}
Under PGD$^{20}$ attacks, SONet with natural training achieves better robust accuracy than TRADES-adversarial training, without sacrificing natural accuracy. The robust accuracy against 20-step PGD attack is a common metric for evaluating $\ell_{\infty}$ adversarial robustness~\cite{madry2018towards, zhang2019theoretically}. We summarize the natural accuracy $\cA_{\nat}$ and robust accuracy $\cA_{\adv}$ under PGD adversarial attacks on different models in Table~\ref{table:postive_result}, where we use PGD$^{k}_{\ast}$ to denote the $k$-step iterative PGD attack within $\ast$-norm box. The natural accuracy of SONet with 32-channel and 64-channel are $88.08\%$ and $89.36\%$ respectively, significantly better than that of ResNet10 with 32-channel and 64-channel deteriorates as $81.52\%$ and $82.74\%$ trained by TRADES ($1/\lambda=1.0$). Under PGD$^{20}_\infty$ attack, our proposed SONet with 64-channel can achieve $61.62\%$ robust accuracy, which is significantly better than the corresponding ResNet10 model with TRADES training (both $1/\lambda=1.0$ and $1/\lambda=6.0$). 

We also evaluate both models against PGD$^{20}_2$ ($\ell_2$-norm $\epsilon=0.5$) adversarial attack. We can observe that SONet is robust against PGD$^{20}_2$ attack, and achieves better robust accuracy than ResNet10 trained by TRADES.

\begin{table*}[t]
\small
\renewcommand\arraystretch{1.2}
\caption{Comparisons between SONet, SOBlock with natural training and ResNet10 with TRADES under white-box PGD adversarial attacks on CIFAR10 dataset.
}
\medskip
\centering
\begin{tabular}{c|c|c|c|c|c}
\hline
\multirow{2}{*}{Model} & \multirow{2}{*}{Channel} &
\multirow{2}{*}{\tabincell{c}{Under \\which attack}} & \multirow{2}{*}{$\cA_\nat(f)$} & \multicolumn{2}{c}{$\cA_\adv(f) $} \\\cline{5-6} 
&    &  &  & $\epsilon = 0.031 (\ell_\infty)$    & \multicolumn{1}{c}{$\epsilon = 0.5 (\ell_2)$}
\\ \hline \hline
SONet & 32 & 
PGD$^{20}$& 88.08\% & 53.67\% & \multicolumn{1}{c}{57.39\%}\\ \hline
SOBlock & 32 & PGD$^{20}$& 90.28\% & 58.21\% & \multicolumn{1}{c}{60.25\%}\\ \hline
ResNet10-\small{TRADES ($1/\lambda=1.0$)} & 32 &
PGD$^{20}$  & 81.52\% & 35.26\%& \multicolumn{1}{c}{57.07\%} \\ \hline
ResNet10-\small{TRADES ($1/\lambda=6.0$)} & 32 & 
PGD$^{20}$  & 73.69\% & 43.46\%& \multicolumn{1}{c}{55.73\%} \\ \hline
SONet & 64 &
PGD$^{20}$  & 89.36\% & 61.62\%& \multicolumn{1}{c}{64.08\%} \\ \hline
SOBlock & 64 &
PGD$^{20}$  & \textbf{91.57\%} & \textbf{62.35\%}& \multicolumn{1}{c}{\textbf{64.70\%}} \\ \hline
ResNet10-\small{TRADES ($1/\lambda=1.0$)} & 64 & 
PGD$^{20}$  & 82.74\% & 37.64\%& \multicolumn{1}{c}{58.97\%} \\ \hline
ResNet10-\small{TRADES ($1/\lambda=6.0$)} & 64 & 
PGD$^{20}$  & 76.29\% & 45.24\%& \multicolumn{1}{c}{57.28\%} \\ \hline
\hline
SONet & 32 & 
PGD$^{1,000}$   & 88.08\% & 19.62\%& \multicolumn{1}{c}{31.75\%} \\ \hline
SOBlock & 32 & 
PGD$^{1,000}$   & 90.28\% & 52.01\%& \multicolumn{1}{c}{52.79\%} \\ \hline
ResNet10-\small{TRADES ($1/\lambda=1.0$)} & 32 & 
PGD$^{1,000}$  & 81.52\% & 33.60\%& \multicolumn{1}{c}{{56.70\%}} \\ \hline
ResNet10-\small{TRADES ($1/\lambda=6.0$)} & 32 & 
PGD$^{1,000}$   & 73.69\% & 43.30\%& \multicolumn{1}{c}{55.48\%} \\ \hline
SONet & 64 & 
PGD$^{1,000}$  & 89.36\% & 24.25\%& \multicolumn{1}{c}{39.79\%} \\ \hline
SOBlock & 64 & 
PGD$^{1,000}$  & \textbf{91.57\%} & \textbf{55.43\%}& \multicolumn{1}{c}{57.37\%} \\ \hline
ResNet10-\small{TRADES ($1/\lambda=1.0$)} & 64 & 
PGD$^{1,000}$  & 82.74\% & 35.78\%& \multicolumn{1}{c}{\textbf{58.73\%}} \\ \hline
ResNet10-\small{TRADES ($1/\lambda=6.0$)} & 64 & 
PGD$^{1,000}$  & 76.29\% & 44.70\%& \multicolumn{1}{c}{56.87\%} \\ \hline
\end{tabular}
\label{table:postive_result}
\vskip -0.1in
\end{table*}

\subsubsection{Nontrivial robustness of SONet under PGD$^{1,000}$ attacks }
In addition to the PGD$^{20}$ attack, we also conduct PGD attacks with more attack steps to better approximate the worst-case attacks. The robust accuracy of all the models decreases with more attack steps (1,000 step), especially for $\ell_{\infty}$ attacks. However, SONet can still achieve $24.25\%$ and $39.79\%$ robust accuracy against PGD$^{1,000}_\infty$ and PGD$^{1,000}_2$ attacks, respectively. Such a decay is worse than TRADES-adversarial training that achieves robust accuracy at $44.70\%$ ($1/\lambda=6.0$) and $58.73\%$ ($1/\lambda=1.0$), but is still non-trivial. Therefore, adversarial robustness of SONet deteriorates but is still nontrivial under further iterative attack steps in PGD$^{1000}$. 

However, a better performance can be achieved by SOBlock below.

\subsubsection{Improved robustness of SOBlock at the first layer than full SONet}
A surprising observation is that by only adopting stablized neural ODE block in the first layer of ResNet10, SOBlock achieves even better performance than SONet that using all layers as such blocks. In Table~\ref{table:postive_result}, the natural accuracy of SOBlock with 32-channel and 64-channel are $90.28\%$ and $91.57\%$ respectively while maintains PGD$^{1000}_\infty$ robust accuracy with $52.01\%$ and $55.43\%$ respectively which is much higher than $43.30\%$ and $44.70\%$ achieved by TRADES ($1/\lambda=6.0$). 

SOBlock almost achieves the best performance among nearly in all settings, except for PGD$^{1000}_2$ attack it has a comparative robust accuracy with TRADES. 
In addition to achieving such a high performance in accuracy, SOBlock particularly enjoys much less computational and memory cost than SONet, that is favoured in applications. 

\section{Gradient Masking Effect by Adaptive Stepsize}
In this section, we further explore the fragility of the adversarial robustness of our proposed architecture under the variation of stepsize choices, robust gradient (CW) attacks, and gradient-free (SPSA) attacks. These results suggest that the adversarial robustness of (stablized) ODE block or net under PGD attacks is likely due to that adaptive stepsize in numerical integration has a gradient masking effect. Gradient masking~\cite{papernot2017practical, athalye2018obfuscated} is a phenomenon widely associated with the obfuscation of gradient information in gradient based adversarial attacks, yet failure under robust gradient and gradient-free attacks, thus giving a false sense of adversarial robustness.

\subsection{Robustness of SOBlock as a result from Adaptive Stepsize in Numerical Integration}
To investigate the reason of adversarial robustness of SOBlock under PGD attacks, we conduct an ablation study on the influence of different order of numerical ODE solvers together with their choice of step size and error tolerance. We use WRN-34-10 as our base network for SOBlock in this section in order to obtain better comparison. The experiments below suggest that adversarial robustness of SOBlock comes from gradient masking effect of adaptive stepsize numerial ODE solvers, including adaptive Heun, Bosh3, and DOPRI5. 

\subsubsection{Adversiarial robustness is only associated with adaptive stepsize ODE solvers}
In the first experiment, we compared three different choices of ODE solvers: Euler method (first order, fixed step size $h=1$), RK4 (fourth order, fixed step size $h=1$), adaptive Heun (second order, adaptive step size), Bosh3 (third order, adaptive stepsize) and DOPRI5 (fifth order, adaptive step size, with default error tolerance ${\tt atol}={\tt rtol}={\tt tol}=0.1$). In Table~\ref{table:comp-solvers}, it shows that all adaptive stepsize solvers (Heun, Bosh3, and DOPRI5) lead to adversarial robustness of SOBlock against PGD attacks, while SOBlocks trained by fixed stepsize solvers Euler and RK4 totally fail under both PGD$^{20}$ and PGD$^{1000}$ in spite of high natural accuracy. 

The same phenomenon persists when we change SOBlock to traditional ODENet \cite{chen2018neural} without using the skew-symmetric stabilization in \eqref{equ: stable ODE block system}. Altough ODENet slightly drops the natural accuracy as desired, one can see in Table~\ref{table:comp-solvers} that robust accuracy of ODENet with both Euler and RK4 fixed step size solver training totally fails ($0\%$), while adaptive step size solver like DOPRI5 shows nontrivial robust accuracy under PGD$^{20}$ and PGD$^{1000}$.

Therefore, adversarial robustness of both ODENet and our stabilized ODE block/net is necessarily associated with the adaptive stepsize in numerical integrations, rather than the fixed stepsize.

\begin{table*}[t]
\small
\renewcommand\arraystretch{1.2}
\caption{Comparisons between SOBlock and ODENet with different solver ODE solvers under PGD adversarial attacks on CIFAR10 dataset.
}
\medskip
\centering
\begin{tabular}{c|c|c|c|c}
\hline
\multirow{2}{*}{Model} & \multirow{2}{*}{Solver} & \multirow{2}{*}{$\cA_\nat(f)$} & \multicolumn{2}{c}{$\cA_\adv(f) $} \\\cline{4-5} 
&    &  & PGD$^{20}$    & \multicolumn{1}{c}{PGD$^{1000}$}
\\ \hline \hline
SOBlock & Euler & 
 94.41\% & 0\% & \multicolumn{1}{c}{0\%}\\ \hline
SOBlock & RK4$_{3/8\mbox{ rule}}$ & 
 92.06\% & 0\% & \multicolumn{1}{c}{0\%}\\ \hline
SOBlock & Dopri5(tol=0.1) & 94.22\% & 71.20\% & \multicolumn{1}{c}{63.20\%}\\ \hline
SOBlock & Dopri5(tol=0.01) &
 93.98\% & 64.66\%& \multicolumn{1}{c}{46.40\%} \\ \hline
SOBlock & Dopri5(tol=0.001) &
94.32\% & 63.87\%& \multicolumn{1}{c}{46.20\%} \\ \hline
SOBlock & Bosh3(tol=0.1) & 92.85\% & {66.00\%} & \multicolumn{1}{c}{{52.84\%}}\\ \hline
SOBlock & Bosh3(tol=0.01) &
 92.30\% & 67.06\%& \multicolumn{1}{c}{59.74\%} \\ \hline
SOBlock & Bosh3(tol=0.001) &
92.38\% & 65.03\%& \multicolumn{1}{c}{55.31\%} \\ \hline
SOBlock & Adaptive Heun(tol=0.1) & 92.42\% & {61.16\%} & \multicolumn{1}{c}{{55.84\%}}\\ \hline
SOBlock & Adaptive Heun(tol=0.01) &
 92.43\% & 63.95\%& \multicolumn{1}{c}{53.79\%} \\ \hline
SOBlock & Adaptive Heun(tol=0.001) &
92.73\% & 57.68\%& \multicolumn{1}{c}{45.33\%} \\ \hline
\hline
ODENet & Euler & 
87.04\% & 0\% & \multicolumn{1}{c}{0\%}\\ \hline
ODENet & RK4$_{3/8\mbox{ rule}}$ & 
87.78\% & 0\% & \multicolumn{1}{c}{0\%}\\ \hline
ODENet & Dopri5(tol=0.1) & 87.41\% & 42.69\% & \multicolumn{1}{c}{13.14\%}\\ \hline
ODENet & Dopri5(tol=0.01) &
87.46\% & 37.20\%& \multicolumn{1}{c}{8.36\%} \\ \hline
ODENet & Dopri5(tol=0.001) &
87.54\% & 36.19\%& \multicolumn{1}{c}{7.75\%} \\ \hline

\end{tabular}
\label{table:comp-solvers}
\vskip -0.1in
\end{table*}

\subsubsection{Adaptive step sizes with large error tolerance in DOPRI5 allows gradient masking}
Both RK4 and DOPRI5 are high (fourth or fifth) order numerical ordinary differential equation methods, where DOPRI5 enjoys a simple error estimate for adaptive stepsize choice \cite{dopri}. In DOPRI5, it uses six function evaluations to calculate both fourth- and fifth-order accurate solutions, whose difference is taken as the error estimate of the fourth-order solution. Adaptive stepsize is adopted in DOPRI5 when the error estimate is within the tolerance specified by absolute error and relative error tolerances $({\tt atol},{\tt rtol})$, both set to be ${\tt tol}$ here \cite{haierODE}:
\[  \textsf{err}=\sqrt{\frac{1}{n} \sum_{i=1}^{n}\left(\frac{y_{1 i}-\widehat{y}_{1 i}}{\mathrm{sc}_{i}}\right)^{2}}, \ \ \ \mathrm{sc}_{i}={{\tt atol}_{i}}+\max \left(\left|y_{0 i}\right|,\left|y_{1 i}\right|\right) \cdot {\tt {rtol}}_{i}. \]
Therefore in the second experiment, we investigate the influence of changing error tolerance ${\tt tol}$ in DOPRI5.
The \textsf{err} is then compared to $1$ in order to find an optimal choice, where an order $q=\min(p,\hat{p})$ solver may choose the optimal step size as $h_{\mathrm{opt}}=h \cdot (1/\textsf{err})^{1/(q+1)}$. For example, DOPRI5 has $q=4$, whence $h_{\mathrm{opt}}=h \cdot (1/\textsf{err})^{1/5}$; adaptive Heun has order $q+1=2$, and BOSH3 has order $q+1=3$. Some care is now necessary for a good implementation: the formula above is multiplied by a safety factor $\textsf{safety}$, for example $\textsf{safety}=0.8,0.9,(0.25)^{1 /(q+1)},$ or $(0.38)^{1 /(q+1)},$ so that the error will be acceptable the next time with high probability. Further, $h$ is not allowed to increase nor to decrease too fast. For example, we may put
$$ h_{\text {new}}=h \cdot \min \left({\tt {ifactor}}, \max \left({\textsf{safety}} \cdot(1 / \textsf{err})^{1 /(q+1)},  {\tt {dfactor}}\right)\right) $$
for the new step size. Then, if $\textsf{err} \leq 1,$ the computed step is accepted and the solution is advanced with $y_{1}$ and a new step is tried with $h_{\text{new }}$ as step size. Else, the step is rejected and the computations are repeated with the new step size $h_{\text {new }}$. The maximal step size increase ${\tt {ifactor}}=10.0$ and the minimial step size decrease ${\tt {dfactor}}=0.2$ by default.

In Table~\ref{table:comp-solvers}, one can see that reasonably large error tolerance in DOPRI5 increases the robustness of both SOBlock and ODENet, e.g. PGD$^{20}_\infty$-robust accuracy at $71.20\%$ at ${\tt tol}=0.1$ against $63.87\%$ at ${\tt tol}=0.001$ for SOBlock and $42.69\%$ at ${\tt tol}=0.1$ against $36.19\%$ at ${\tt tol}=0.001$ for ODENet. Large error tolerance leads to large perturbations on adaptive stepsize in DOPRI5, e.g. Table~\ref{table:comp-steps} shows that increasing tolerance from ${\tt tol}=0.001$ to ${\tt tol}=0.1$ lead to enlarged adaptive stepsize perturbations from the order of $1e-3$ to $1e-2$.  

These phenomena above show that adversarial robustness under PGD attacks is a result from the adaptive stepsize choice of numerical ODE solvers that perturbs the gradients of loss functions, where enlarging error tolerance properly may increase the robustness of SOBlocks and ODENets. Therefore, adaptive step size ODE solvers like DOPRI5 contribute such a kind of gradient masking against PGD attacks: reasonably large error tolerance in numerical function estimate leads to large perturbations of gradients that fools the projected gradient descent in attacks. We also note that over-enlarging error tolerance, especially in low order ODE solvers (adaptive Heun and Bosh3), may lead to inaccuracies in natural training that eventually drops robust accuracy as well. Hence one should expect a reasonable choice of error tolerance should depend on a trade-off between fitting accuracy and gradient masking effect. 

\begin{table*}[t]
\renewcommand\arraystretch{1.2}
\caption{Adaptive steps with ODENet under different dopri5 tolerances and PGD$_\infty$ attack iterations}
\medskip
\centering
\begin{tabular}{c|c|c}
\hline
Solver & PGD iterations & Adaptive steps\\\hline \hline
\multirow{3}*{Dopri5(tol=0.1)} & 1 & [0.0, 0.262, 1.0]\\ \cline{2-3}
~ & 100 & [0.0, 0.253, 1.0]\\ \cline{2-3}
~ & 1000 & [0.0, 0.244, 1.0]\\ \hline \hline
\multirow{3}*{Dopri5(tol=0.01)} & 1 & [0.0, 0.155, 0.827, 1.0]\\ \cline{2-3}
~ & 100 & [0.0, 0.150, 0.793, 1.0]\\ \cline{2-3}
~ & 1000 & [0.0, 0.149, 0.789, 1.0]\\ \hline \hline
\multirow{3}*{Dopri5(tol=0.001)} & 1 & [0.0, 0.097, 0.423, 1.0]\\ \cline{2-3}
~ & 100 & [0.0, 0.096, 0.420, 1.0]\\ \cline{2-3}
~ & 1000 & [0.0, 0.094, 0.409, 0.981, 1.0]\\ \hline

\end{tabular}
\label{table:comp-steps}
\vskip -0.1in
\end{table*}


\subsection{Robust Gradient (CW) and Gradient-Free (SPSA) Attacks}
To further justify our reasoning above that the adversarial robustness of SOBlock and SONet is due to the gradient masking effect of adaptive stepsize numerical ODE solvers, especially DOPRI5, we further conduct experiments under two sorts of new attacks, CW attack that has robust gradients due to the use of hinge loss and SPSA attack that is a kind of gradient-free attack. 

Table~\ref{table:neg-result} shows that in spite of the impressive robustness under PGD attacks, both SONet and SOBlock are vulnerable under CW$_\infty$ and SPSA attacks, while ResNet10 trained with TRADES still has relatively strong robustness. Particularly, under CW$_\infty$ attack, SOBlock (SONet) with 64 channels has $0\%$ ($11.20\%$) robust accuracy compared with ResNet10 in TRADES training at $39.77\%$ ($1/\lambda=6$); while under SPSA attack, SOBlock (SONet) of 64 channels has $11.68\%$ ($15.10\%$) robust accuracy compared with TRADES training at $69.97\%$ ($1/\lambda=1$). This provides a support of the gradient masking by DOPRI5, that fails to fool CW$_\infty$ and SPSA attacks which are not as sensitive to gradients of cross entropy loss as PGD attacks. Hence the gradient masking of adaptive stepsize gives us a false sense of adversarial robustness in PGD attacks. 


\begin{table*}[t]
\small
\renewcommand\arraystretch{1.2}
\caption{Comparisons between SONet, SOBlock with natural training and ResNet10 with TRADES under CW$_\infty$ and SPSA adversarial attacks on CIFAR10 dataset.
}
\medskip
\centering
\begin{tabular}{c|c|c|c|c}
\hline
\multirow{2}{*}{Model} & \multirow{2}{*}{Channel} & \multirow{2}{*}{$\cA_\nat(f)$} & \multicolumn{2}{c}{$\cA_\adv(f) $} \\\cline{4-5} 
&    &  & CW-Linf    & \multicolumn{1}{c}{SPSA}
\\ \hline \hline
SONet & 32 & 
 88.08\% & 0\% & \multicolumn{1}{c}{2.50\%}\\ \hline
SOBlock & 32 & \textbf{90.28\%} & 0\% & \multicolumn{1}{c}{7.64\%}\\ \hline
ResNet10-\small{TRADES ($1/\lambda=1$)} & 32 & 
 81.52\% & 37.61\%& \multicolumn{1}{c}{\textbf{68.30\%}} \\ \hline
ResNet10-\small{TRADES ($1/\lambda=6$)} & 32 &
 73.69\% & {\bf 38.92}\%& \multicolumn{1}{c}{63.60\%} \\ \hline
SONet & 64 &
 89.36\% & 11.20\%& \multicolumn{1}{c}{15.10\%} \\ \hline
SOBlock & 64 &
\textbf{91.57\%} & 0\%& \multicolumn{1}{c}{11.68\%} \\ \hline
ResNet10-\small{TRADES ($1/\lambda=1$)} & 64 & 
 82.74\% & 35.78\%& \multicolumn{1}{c}{\textbf{69.97\%}} \\ \hline
ResNet10-\small{TRADES ($1/\lambda=6$)} & 64 & 
 76.29\% & {\bf 39.77}\%& \multicolumn{1}{c}{65.97\%} \\ \hline
\end{tabular}
\label{table:neg-result}
\vskip -0.1in
\end{table*}

\section{Conclusions}
In this paper, we propose a stabilized neural ODE architecture based on a skew-symmetric dynamical system with provable Lyapunov stability. We show that such an ODE based network architecture can achieve some state-of-the-art adversarial robustness with natural training against PGD attacks, without sacrificing natural accuracy that is suffered by popular adversarial training methods such as TRADES. To understand this phenomenon, we further explore the possible mechanism underlying such kind of adversarial robustness. We show that the adaptive stepsize numerial ODE solvers, such as adaptive HEUN2, BOSH3, and especially DOPRI5, have a gradient masking effect that fails the PGD attacks which are sensitive to gradient information of training loss, while they can not fool the CW attack of robust gradients and the SPSA attack that is gradient-free. This provides a new explanation that the adversarial robustness of ODE based networks is mainly due to the obfuscated gradients in numerical ODE solvers with adaptive step sizes. 

\section*{Acknowledgments}
This research made use of the computing resources of the X-GPU cluster supported by the Hong Kong Research Grant Council Collaborative Research Fund: C6021-19EF. The research of Yifei Huang and Yuan Yao is supported in part by HKRGC 16303817, ITF UIM/390, as well as awards from Tencent AI Lab and Si Family Foundation. Yaodong Yu and Yi Ma acknowledge support from ONR grant N00014-20-1-2002 and the joint Simons Foundation-NSFDMS grant \#2031899. Hongyang Zhang was supported in part by the Defense Advanced Research Projects Agency under cooperative agreement HR00112020003. The views expressed in this work do not necessarily reflect the position or the policy of the Government and no official endorsement should be inferred. Approved for public release; distribution is unlimited.

\newpage
\bibliographystyle{alpha}
\bibliography{reference}

\end{document}